\documentclass{article}

\usepackage[letterpaper,margin=0.85in]{geometry}
\usepackage{titling}
\usepackage{authblk}

\usepackage[utf8]{inputenc}
\usepackage[T1]{fontenc}
\usepackage{hyperref}
\usepackage{url}
\usepackage{booktabs}
\usepackage{amsfonts}
\usepackage{nicefrac}
\usepackage{microtype}
\usepackage{xcolor}
\usepackage{graphicx}
\usepackage{subcaption}
\usepackage{tabularx}
\usepackage{siunitx}
\usepackage{multirow}
\usepackage{threeparttable}
\usepackage{placeins}
\usepackage{tikz}
\usepackage{listings}
\usepackage[most]{tcolorbox}
\usepackage{caption}
\usepackage{algorithm}
\usepackage{algpseudocode}
\usepackage{float}
\usepackage{amsmath}
\usepackage{amssymb}
\usepackage{mathtools}
\usepackage{amsthm}
\usepackage{enumitem}
\usepackage{pifont}
\usepackage[capitalize,noabbrev]{cleveref}

\pretitle{\begin{center}\LARGE\bfseries\hrule height 1pt \vspace{0.6em}}
\posttitle{\vspace{0.6em}\hrule height 1pt\end{center}}
\preauthor{\begin{center}\large}
\postauthor{\end{center}}
\predate{\begin{center}\small}
\postdate{\end{center}}

\algnewcommand\algorithmicoutput{\textbf{Output:}}
\algnewcommand\Output{\item[\algorithmicoutput]}
\usetikzlibrary{arrows.meta,positioning}

\lstdefinestyle{dictstyle}{
  basicstyle=\ttfamily\footnotesize,
  columns=fullflexible,
  keepspaces=true,
  breaklines=true,
  breakatwhitespace=false,
  showstringspaces=false,
  frame=single,
  rulecolor=\color{black!20},
  backgroundcolor=\color{black!2},
  xleftmargin=0.5em,
  framexleftmargin=0.5em,
  tabsize=2
}

\newtcolorbox{promptbox}[2][]{%
  enhanced,
  breakable,
  colback=blue!5,
  colframe=blue!70!black,
  coltitle=white,
  fonttitle=\bfseries,
  title=#2,
  arc=4mm,
  boxrule=0.8pt,
  boxsep=6pt,
  before skip=8pt,
  after skip=8pt,
  #1
}

\newcommand{\heatcell}[4]{%
  \begingroup
  \pgfmathsetmacro{\t}{(#1-#2)/(#3-#2)}%
  \pgfmathsetmacro{\x}{max(0,min(1,\t))}%
  \pgfmathsetmacro{\rR}{0.971} \pgfmathsetmacro{\gR}{0.746} \pgfmathsetmacro{\bR}{0.746}%
  \pgfmathsetmacro{\rG}{0.55}  \pgfmathsetmacro{\gG}{0.95}  \pgfmathsetmacro{\bG}{0.65}%
  \pgfmathsetmacro{\u}{min(1,max(0,2*\x))}%
  \pgfmathsetmacro{\v}{min(1,max(0,2*\x-1))}%
  \pgfmathsetmacro{\rr}{(1-\u)*\rR + \u*1}%
  \pgfmathsetmacro{\gr}{(1-\u)*\gR + \u*1}%
  \pgfmathsetmacro{\br}{(1-\u)*\bR + \u*1}%
  \pgfmathsetmacro{\r}{(1-\v)*\rr + \v*\rG}%
  \pgfmathsetmacro{\g}{(1-\v)*\gr + \v*\gG}%
  \pgfmathsetmacro{\b}{(1-\v)*\br + \v*\bG}%
  \definecolor{heatcol}{rgb}{\r,\g,\b}%
  \tikz[baseline=(X.base)]\node[fill=heatcol,rounded corners=1.5pt,inner xsep=2pt,inner ysep=1pt](X){#4};%
  \endgroup
}

\theoremstyle{plain}
\newtheorem{theorem}{Theorem}[section]

\theoremstyle{definition}

\theoremstyle{remark}

\newcommand{\cmark}{\ding{51}}
\newcommand{\xmark}{\ding{55}}
\makeatletter

\newcommand{\Rmnum}[1]{\expandafter\@slowromancap\romannumeral #1@}
\renewcommand{\theHALG@line}{\thealgorithm.\arabic{ALG@line}}
\makeatother

\title{AscendOptimizer: Episodic Agent for Ascend NPU Operator Optimization}
\author[1,\thanks{Equal contribution.}]{Jiehao Wu}
\author[1,$^*$]{Zixiao Huang}
\author[2]{Wenhao Li}
\author[3]{Chuyun Shen}
\author[1,\thanks{Corresponding author.}]{Junjie Sheng}
\author[4,5,6,$^{\dagger}$]{Xiangfeng Wang}
\affil[1]{School of Computer Science and Technology, East China Normal University}
\affil[2]{School of Computer Science and Technology, Tongji University}
\affil[3]{Shanghai University of International Business and Economics}
\affil[4]{Key Lab of Mathematics and Engineering Applications (MoE), East China Normal University}
\affil[5]{School of Mathematical Sciences, East China Normal University}
\affil[6]{Shenzhen Loop Area Institute (SLAI)}
\date{}

\begin{document}

\maketitle
\vspace{-4.8em}
\begin{center}
\IfFileExists{fig/logo_name.png}{\includegraphics[width=0.22\textwidth]{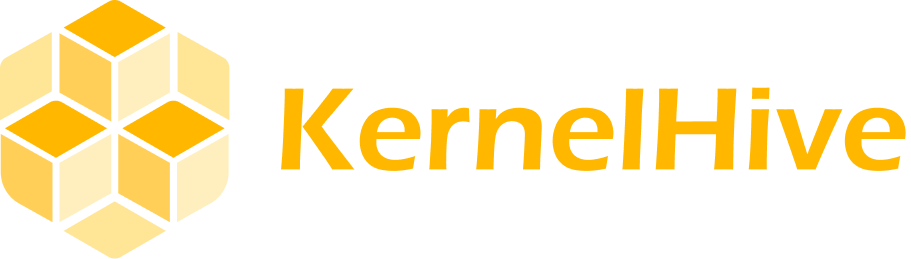}\\[-0.1em]}{}
Project Website: \href{https://github.com/KernelHive}{\textcolor{magenta}{https://github.com/KernelHive}}
\end{center}

\begin{abstract}
Optimizing AscendC (Ascend C) operators for Ascend NPUs is difficult for two reasons. First, unlike CUDA, the ecosystem offers few public kernels to learn from. Second, performance depends on a coupled two-part implementation: a host-side tiling program that controls data movement and a kernel program that schedules and pipelines computation. We present \textbf{AscendOptimizer}, an episodic agent that builds missing optimization knowledge from execution itself.
For kernel optimization, AscendOptimizer \emph{rewinds} strong implementations by removing optimizations in a controlled way, then keeps the changes whose removal measurably hurts performance as reusable experience for later rewriting. For host-side optimization, it runs profiling-in-the-loop evolutionary search to find valid, fast tiling and data-movement configurations directly from hardware feedback. This combination lets the agent improve kernel structure and host-side scheduling together.
On a benchmark of 101 real AscendC operators, AscendOptimizer achieves a \num{1.21}$\times$ geometric-mean speedup over the open-source baseline, and \SI{53.47}{\percent} of operators run faster than their references. Given a same budget of evaluations per operator, AscendOptimizer consistently outperforms Best-of-$N$ sampling and OpenEvolve in terms of geometric mean speedup, $\text{fast}_p$ tail speedup ratios, and overall optimization progress across varying budgets.

\end{abstract}

\section{Introduction}
\label{sec:introduction}

LLM-based kernel optimization has recently shown strong results on platforms with rich
open-source ecosystems, especially CUDA and Triton~\cite{ouyang2025kernelbench,li2025tritonbench}.
These successes are partially enabled by the mature CUDA/Triton ecosystem. 
Modern code LMs are often trained on large public-code corpora, and prior work shows that such models can memorize or reproduce code fragments from training data~\cite{chen2021evaluating,kocetkov2022stack, yang2024unveiling}. Meanwhile, CUDA/Triton provide many public optimized implementations and reusable performance idioms, including tiling, hierarchical data movement, and memory-efficient fused kernels~\cite{tillet2019triton,cutlass,dao2022flashattention}. 
As a result, LLM-based optimization agents can benefit not only from reasoning about programs, but also from adapting implementation patterns that are already visible in the ecosystem.
On emerging or restricted-access hardware, this support is much weaker. Prior studies report sharply lower LLM kernel-generation success on AscendC than on CUDA, and attribute this gap in part to limited representation of AscendC-specific syntax, execution models, and optimization principles in public or pretraining corpora~\cite{wen2025multikernelbench,cao2026ascendkernelgen}. 
Platform-specific optimization patterns are therefore less accessible to current models, and compiler or profiler feedback often identifies symptoms without directly specifying the structural rewrite needed to fix them.
In this setting, relying solely on pretrained code knowledge or one-shot generation is insufficient,
which raises a central question: \emph{when external references are limited, how can an optimization agent build the missing expertise for itself?}

A broader family of methods addresses limited supervision by constructing learning
signals from interaction or from the artifact itself. In program repair, self-supervised
approaches such as BugLab~\cite{allamanis2021self} and SelfAPR~\cite{ye2022selfapr}
synthesize buggy programs from correct code and learn to recover the original program
using compiler or test feedback. In agent settings, Reflexion~\cite{shinn2023reflexion}
externalizes trial-and-error feedback into episodic memory, while Voyager~\cite{wang2023voyager}
stores reusable executable skills in a growing skill library. Hindsight-style reinforcement
learning similarly shows that sparse feedback can be made useful by reinterpreting
trajectories after the fact~\cite{andrychowicz2017hindsight}. These works suggest that, when external demonstrations are scarce, useful supervision can be constructed from the artifact itself or from interaction feedback. Following this principle, we treat optimized kernels as latent sources of optimization knowledge: a code motif whose removal measurably hurts hardware performance becomes evidence of a reusable optimization pattern.

We instantiate this perspective as \textbf{Optimization Rewind}: instead of
assuming exact reversibility, we construct semantically valid de-optimized
neighbors of strong kernels by removing recognizable optimization motifs, and
retain only removals that pass validation but measurably hurt hardware
performance. Each validated degradation becomes evidence that the removed motif
was useful in its original context, and is distilled into reusable
forward-optimization experience for later code rewriting. This mechanism is
related to prior de-optimization~\cite{hines2005using} and rewind-style
self-supervision~\cite{zhang2025rewind}, but is used here to mine practical
optimization knowledge under scarce external demonstrations.

We study this problem on Ascend NPUs, which provide a particularly strict testbed for knowledge-scarce optimization. Recent MultiKernelBench results~\cite{wen2025multikernelbench} show a large generalization gap: while state-of-the-art models reach \SIrange{44.2}{52.6}{\percent} Pass@1 for one-shot CUDA operator generation, the AscendC pass rate stays below \SI{2.1}{\percent}. As Table~\ref{tab:pass_rate_gap} shows, this gap is not just a syntax problem. AscendC exposes an explicitly managed memory hierarchy and requires developers to orchestrate data movement and synchronization through the on-chip Unified Buffer (UB)~\cite{zhou2025squeezing}. An AscendC operator is not a single kernel, it is a \emph{two-part artifact}: a host-side \emph{tiling} program decides how data are partitioned and moved, and a device-side \emph{kernel} program decides how computation is scheduled and pipelined. Performance therefore depends on both \emph{how data are fed} and \emph{how the kernel runs}, which makes Ascend a strong example of coupled optimization under knowledge scarcity.

\begin{table}[htbp]
\centering
\caption{One-shot operator generation pass rate (Pass@1) across hardware platforms, reported by MultiKernelBench~\cite{wen2025multikernelbench}. The results confirm severe knowledge scarcity on Ascend.}
\label{tab:pass_rate_gap}
{\small
\renewcommand{\arraystretch}{0.9}
\begin{tabular}{lcc}
\toprule
\textbf{Model} & \textbf{CUDA (Pass@1)} & \textbf{AscendC (Pass@1)} \\
\midrule
DeepSeek-R1 & \SI{52.6}{\percent} & \SI{1.4}{\percent}  \\
Claude-Sonnet-4 & \SI{47.0}{\percent} & \SI{2.1}{\percent}  \\
Qwen3-235B (think) & \SI{44.2}{\percent} & \SI{0.7}{\percent}  \\
\bottomrule
\end{tabular}}%
\end{table}

Motivated by this setting, we propose \textbf{AscendOptimizer}, a two-stage optimization framework for knowledge-scarce operator development. The two stages target different optimization handles. Stage~\Rmnum{1} focuses on kernel refinement: it runs benchmark-wide Optimization Rewind over the strongest available implementations, distills the validated trajectories into a shared experience bank, and then uses retrieval-augmented rewriting to guide later kernel edits under compiler and profiling feedback. This stage targets the part that sparse feedback alone struggles to teach: \emph{how the code should be rewritten}. Stage~\Rmnum{2} addresses a different problem on the host side. The tiling space is highly discontinuous, and small changes in tile sizes or data-movement schedules can turn a valid fast configuration into an invalid one. Instead of trying to encode these rules in advance, AscendOptimizer uses \textbf{evolution-guided program search} with hardware-in-the-loop feedback to discover strong tiling configurations directly from execution results. The two stages are complementary but asymmetric: Stage~\Rmnum{1} provides the main structural gains inside the kernel, while Stage~\Rmnum{2} adds host-side tiling and resource-policy improvements once the refined kernel exposes better execution opportunities.

Our contributions are threefold:
\textbf{1)} We introduce \textbf{Optimization Rewind}, a self-supervised way to bootstrap kernel optimization experience under knowledge scarcity by degrading strong implementations in a controlled manner and distilling the removals that verifiably matter for performance.
\textbf{2)} We formulate AscendC optimization as a coupled problem of \emph{host-side tiling} and \emph{device-side kernel refinement}, and instantiate this view in \textbf{AscendOptimizer}, which combines retrieval-guided kernel rewriting with hardware-guided tiling search.
\textbf{3)} On \textbf{101} real AscendC operators, AscendOptimizer achieves a \num{1.21}$\times$ geometric-mean speedup over the open-source baseline, showing that this approach is effective in a realistic knowledge-scarce setting.

\section{Related Work}
\label{sec:related-work}

\noindent\textbf{Traditional Operator Compilation and Domain-Specific Architecture Optimization.}
Operator systems such as TVM~\cite{chen2018tvm}, Halide~\cite{ragan2013halide}, Triton~\cite{tillet2019triton}, TileLang~\cite{wang2025tilelang}, polyhedral compilers~\cite{bondhugula2008pluto,baghdadi2019tiramisu,zhao2021akg}, and search-based compilers~\cite{zheng2020ansor,zheng2021tenset,wu2025mirage} reduce manual tuning effort through DSLs, IRs, cost models, and schedule search. Yet high-end kernels such as FlashAttention~\cite{dao2022flashattention,shah2024flashattention} still require architecture-specific design, and this gap is sharper on Ascend NPUs, whose memory hierarchy and instruction model make optimization heavily dependent on hardware expertise~\cite{zhou2025squeezing,ai2025neutronascend}. Existing Ascend analyses and profilers provide useful signals~\cite{zhou2025accelerating,moustafa2023accelerating}, but still rely on human interpretation.

\noindent\textbf{LLM Agent-based Operator Generation and Iterative Optimization.}
KernelBench~\cite{ouyang2025kernelbench}, TritonBench~\cite{li2025tritonbench}, Astra~\cite{wei2025astra}, PRAGMA~\cite{lei2025pragma}, CudaForge~\cite{zhang2025cudaforge}, KernelEvolve~\cite{liao2025kernelevolve}, Geak~\cite{wang2025geak}, EvoEngineer~\cite{guo2025evoengineer}, TritonForge~\cite{li2025tritonforge}, GPU Kernel Scientist~\cite{andrews2025gpu}, and StitchCUDA~\cite{li2026stitchcuda} show that LLM agents can generate and refine CUDA or Triton kernels using compiler, profiler, multi-agent, and hardware-in-the-loop feedback. Their success, however, is tied to ecosystems where public kernels and optimization idioms are abundant. Direct transfer to DSAs is difficult because aligned corpora and accessible architectural knowledge are scarce~\cite{wen2025multikernelbench,cao2026ascendkernelgen}.

\noindent\textbf{Internalized Optimization through Training.}
Kevin~\cite{baronio2025kevin}, TritonRL~\cite{woo2025tritonrl}, AutoTriton~\cite{li2025autotriton}, CUDA-L1/L2~\cite{li2025cuda,su2025cuda}, and Seed-Coder~\cite{seed2025seed} instead store optimization experience in model parameters through RL, SFT, or large-scale contrastive sampling. These approaches can be effective, but they require many domain-specific code-performance pairs and substantial training cost. In immature hardware ecosystems, that data dependency is often the bottleneck~\cite{wen2025multikernelbench}; AscendOptimizer therefore targets a training-free setting.

\noindent\textbf{LLM-driven NPU Code Generation.}
AscendKernelGen~\cite{cao2026ascendkernelgen}, AscendCraft~\cite{wen2025ascendcraft}, and MultiKernelBench~\cite{wen2025multikernelbench} directly study Ascend or NPU code generation, mainly from high-level specifications or DSLs. AscendOptimizer addresses the complementary problem of improving \emph{existing} functionally correct AscendC implementations, jointly optimizing host-side tiling and device-side kernels while building missing domain knowledge from execution feedback.

\section{The AscendOptimizer Agent}
\label{sec:method}

\begin{figure*}[ht]
\centering
\includegraphics[width=1.0\textwidth]{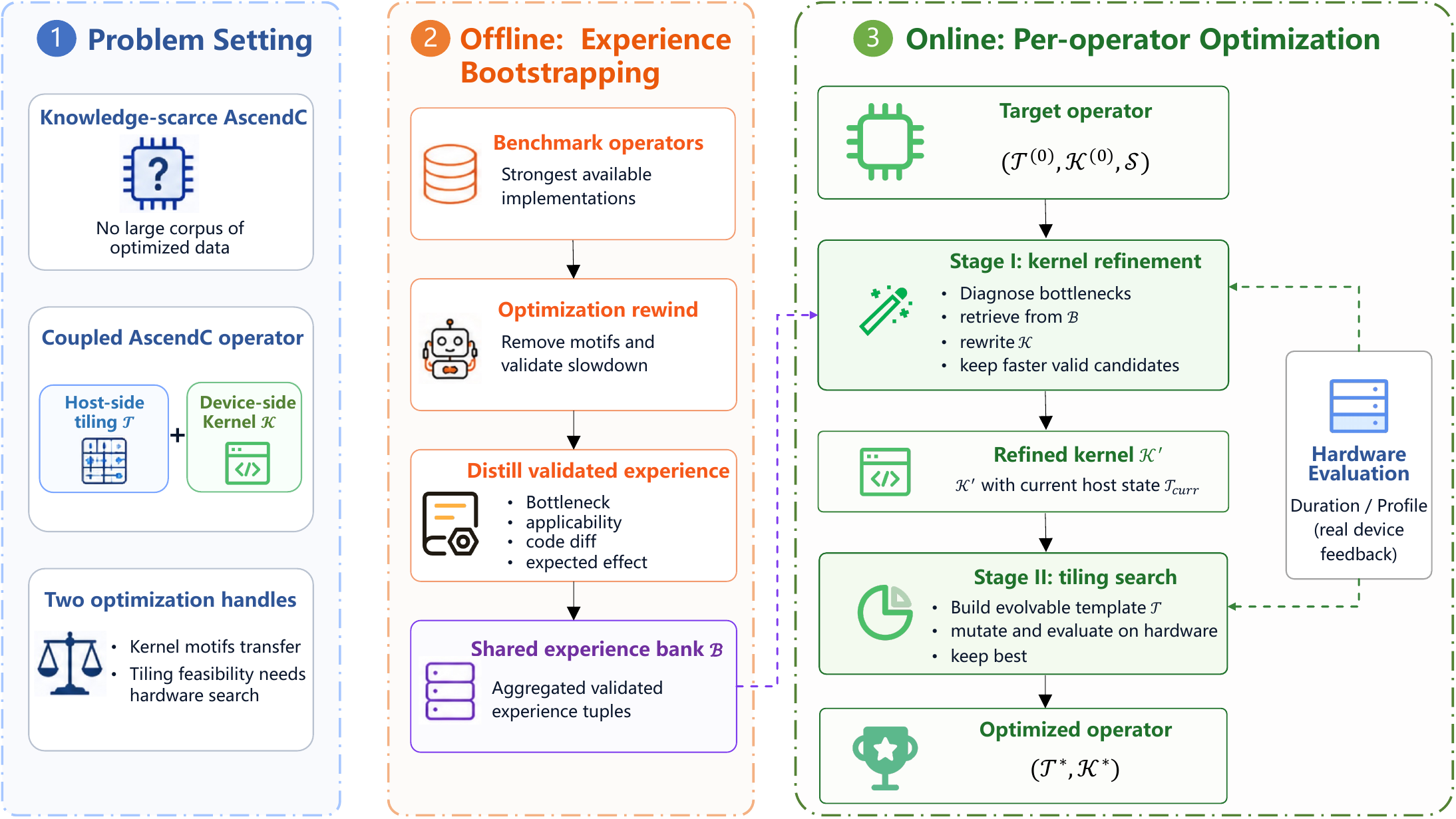}
\caption{Overview of \textbf{AscendOptimizer}. Offline, we rewind benchmark operators and distill validated optimization motifs into a shared experience bank. Online, the bank guides Stage~\Rmnum{1} kernel refinement, after which Stage~\Rmnum{2} searches host-side tiling configurations under the refined kernel.}
\label{framework}
\end{figure*}

\subsection{Method Overview}

Knowledge scarcity does not affect every part of AscendC optimization in the same way. Kernel code contains structured motifs such as pipelining, vectorization, buffer management, tail handling, and synchronization elimination. Because these motifs are compositional, they can be mined from strong existing implementations. Host-side tiling is different. It is tightly constrained by input shapes, alignment rules, buffer capacity, and resource-allocation decisions, so its feasible region is discrete and highly discontinuous. As a result, cross-operator transfer is much weaker for tiling than for kernel structure. We therefore treat the two parts differently instead of forcing them into a single optimizer.

Concretely, \textbf{AscendOptimizer} has a benchmark-wide offline bootstrapping phase and a per-operator online optimization phase. Offline, each operator is explored with \textbf{Optimization Rewind} under a fixed rewind budget. After rewind finishes across the benchmark, the hardware-validated de-optimization pairs are distilled into a shared experience bank. Online, Stage~\Rmnum{1} uses this bank to refine the target kernel, and Stage~\Rmnum{2} then searches host-side tiling configurations under the refined kernel.

\subsection{Problem Formulation}

Under the Ascend NPU heterogeneous computing architecture, we represent the operator to be optimized as a tuple $\mathcal{O} = \langle \mathcal{T}, \mathcal{K}, \mathcal{S} \rangle$, where:
\begin{itemize}[leftmargin=*]
    \item $\mathcal{T} \in \mathbb{C}_{\text{tiling}}$: the host-side tiling function that determines data partitioning, movement, and resource policies.
    \item $\mathcal{K} \in \mathbb{C}_{\text{kernel}}$: the device-side kernel code that determines instruction scheduling, buffering, and synchronization structure.
    \item $\mathcal{S}$: the static operator attributes (e.g., input shape, data type, layout).
\end{itemize}

Given hardware constraints $H$ (e.g., buffer capacity, pipeline stages, and alignment rules), our objective is to identify the tiling function $\mathcal{T}^{*}$ and kernel implementation $\mathcal{K}^{*}$ that minimize end-to-end latency on real hardware:
\begin{equation}
    (\mathcal{T}^*, \mathcal{K}^*) = \mathop{\arg\min}_{\mathcal{T}, \mathcal{K}} \mathcal{L}\left(\text{Exec}(\mathcal{T}, \mathcal{K}, \mathcal{S}) \mid H\right).
    \label{eq:problem_objective}
\end{equation}
Here, $\text{Exec}(\cdot)$ denotes hardware compilation and execution, and $\mathcal{L}$ denotes the measured latency. For brevity, when $H$ and $\mathcal{S}$ are fixed, we write $\mathcal{L}(\mathcal{K}\mid\mathcal{T}_{\text{curr}})$ as shorthand for $\mathcal{L}\!\left(\text{Exec}(\mathcal{T}_{\text{curr}}, \mathcal{K}, \mathcal{S}) \mid H\right)$ and omit $\text{Exec}(\cdot)$ and $H$.

Although $\mathcal{T}$ and $\mathcal{K}$ are coupled, they play different roles. Kernel rewrites change instruction scheduling, buffer allocation, and synchronization topology, which in turn changes which tiling strategies are feasible or worthwhile. Tiling, by contrast, searches for better data partitioning and resource policies under a fixed kernel structure. In our implementation, kernel refinement is evaluated under the current tiling function $\mathcal{T}_{\text{curr}}$, yielding $\mathcal{K}_{\text{curr}}$, and tiling search is then run under that refined kernel. Since $H$ contains many non-differentiable black-box constraints (e.g., bank conflicts, alignment failures, and cache thrashing), and the code space $\mathbb{C}_{\text{tiling}} \times \mathbb{C}_{\text{kernel}}$ is highly discrete and non-convex, direct gradient-based optimization is not applicable.

\subsection{Offline Experience Bootstrapping via Optimization Rewind}

\noindent\textbf{Why Rewind Works.}
High-performance kernels contain layered optimization motifs such as pipelining,
vectorization, buffer management, tail handling, and barrier elimination. These
motifs interact, so Optimization Rewind does not assume exact reversibility.
Instead, many motifs leave recognizable local signatures in code or profiles. We
therefore remove one targeted motif at a time, keep only semantic-preserving
candidates that become measurably slower on hardware, and distill the removed
motif as reusable forward-optimization experience. We formalize this view as a
local one-factor-at-a-time screening on the discrete motif space in
Appendix~\ref{app:rewind_formalization}.

\begin{theorem}[Rewind acceptance is a hardware-validated screening signal]
\label{thm:rewind_screening}
Let $\mathcal{K}_{A}$ be a kernel whose active motif set is $A\subseteq\Omega$,
and let $r_m$ denote the rewind action for motif $m\in A$ under the fixed
tiling $\mathcal{T}$. Define the local one-factor-at-a-time effect
$\widehat{\Delta}_m(\mathcal{K}_{A};\mathcal{T})
=\mathcal{L}(\mathcal{K}_{A\setminus\{m\}}\mid\mathcal{T})
-\mathcal{L}(\mathcal{K}_{A}\mid\mathcal{T})$.
If $r_m(\mathcal{K}_{A})\neq\bot$ (i.e., the de-optimized candidate compiles
and is semantically equivalent) and
$\widehat{\Delta}_m(\mathcal{K}_{A};\mathcal{T})>\tau_{\mathrm{noise}}$, then
removing motif $m$ at basepoint $\mathcal{K}_{A}$ degrades measured latency by
strictly more than profiling noise; equivalently, motif $m$ contributes
positively to performance at $\mathcal{K}_{A}$ and the inverse rewrite
$\mathcal{K}_{A\setminus\{m\}}\!\rightarrow\!\mathcal{K}_{A}$ constitutes a
hardware-validated optimization step.
\end{theorem}

\noindent\textbf{Per-Operator Rewind Trajectory Construction.}
Before any online optimization, we run a benchmark-wide offline bootstrapping phase. For each operator $o$ in the benchmark, we start from its strongest available implementation and allocate a rewind budget $B_r$. In each round, an inverse agent proposes a \emph{semantically meaningful} de-optimization, such as removing double buffering, breaking vectorized data paths, or reintroducing unnecessary synchronization, rather than corrupting the code at random. This produces a validated de-optimization trajectory
\begin{equation}
    \mathbb{T}_{o} = \left(\mathcal{K}^{(0)}_{o}, \mathcal{K}^{(1)}_{o}, \dots, \mathcal{K}^{(T_{o})}_{o}\right), \quad T_{o} \le B_r.
\end{equation}

\noindent\textbf{Hardware-Validated Experience Distillation.}
Each rewound candidate is then subjected to compilation, correctness, and profiling checks on Ascend hardware. We only keep pairs whose latency meaningfully degrades after the semantic de-optimization, which tells us that the removed motif was actually helping performance rather than merely changing style. For each validated pair, the agent distills a structured experience tuple
\begin{equation}
    \mathcal{M} = \langle \text{Title, Bottleneck, Applicability, Expected Effect, Code Diff} \rangle.
    \label{eq:experience_tuple}
\end{equation}
Here, \textit{Bottleneck} and \textit{Applicability} describe when the pattern should be retrieved, while \textit{Expected Effect} and \textit{Code Diff} describe what change should be made and why it should help. After all operators finish their rewind phase, the validated tuples are aggregated into a shared hardware-validated experience bank $\mathcal{B}$.

\subsection{Online Retrieval-Guided Kernel Refinement}

Once the shared bank has been built, each target operator enters online optimization. The offline rewind phase and the online kernel-refinement phase together make up the kernel-side path of Stage~\Rmnum{1}.

\noindent\textbf{Bottleneck Diagnosis.}
Given the current operator state $(\mathcal{T}_{\text{curr}}, \mathcal{K}_{\text{curr}}, \mathcal{S})$, the agent analyzes compiler diagnostics and profiling traces, then summarizes the main structural bottleneck as a query $q$.

\noindent\textbf{Experience Retrieval.}
Using $q$, a dense retriever fetches the Top-$k$ tuples $\{\mathcal{M}_{i}\}_{i=1}^{k}$ from the shared experience bank $\mathcal{B}$. Retrieval depends on bottleneck symptoms and structural applicability, not on operator identity.

\noindent\textbf{Experience-Guided Rewriting and Selection.}
A refiner LLM rewrites $\mathcal{K}_{\text{curr}}$ using the retrieved tuples as structured guidance. Each rewritten candidate is compiled, checked for numerical correctness, and profiled under the current tiling function $\mathcal{T}_{\text{curr}}$. We only accept candidates that both pass validation and improve measured latency, updating $\mathcal{K}_{\text{curr}}$ accordingly. Stage~\Rmnum{1} is therefore more than plain RAG and more than blind trial-and-error: it combines \emph{bootstrapped shared experience} with \emph{hardware-grounded candidate selection}.

\subsection{Hardware-in-the-Loop Tiling Search}

Stage~\Rmnum{2} targets a different optimization object. Unlike kernel motifs, tiling validity depends much more heavily on shape-specific constraints, alignment, buffer capacity, and hardware resource interactions. A small edit can turn a valid fast candidate into an invalid one. For that reason, retrieval is not the main mechanism here. Instead, we rely on hardware-in-the-loop search under the current refined kernel.

\noindent\textbf{Search Space Construction.}
Starting from the original host-side implementation and attributes $\mathcal{S}$, the agent synthesizes an evolvable template $\mathcal{T}_{\text{base}}$ containing mutation markers (Appendix~\ref{app:methodDet}, Figure~\ref{fig:appendix_tiling_code}). The template exposes the locations where tile size, core allocation, buffer policy, and lightweight control logic can be changed.

\noindent\textbf{Mutation.}
At each search step, the LLM mutates the current tiling candidate. Mutations include changes to tile sizes, core counts, buffer policies, and lightweight branch logic for data movement and launch configuration.

\noindent\textbf{Hardware Evaluation and Survivor Update.}
Each mutated tiling candidate is compiled and executed with the current refined kernel $\mathcal{K}_{\text{curr}}$. Candidates that fail compilation or correctness checks are discarded immediately. Among the survivors, the fastest one becomes the parent for the next search step. This keeps the search inside the implicit feasible region defined by hardware constraints while exploiting the execution structure already established by Stage~\Rmnum{1}.

\section{Experiments}
\label{sec:experiments}

We structure the evaluation around four questions: (1) Under a fixed hardware-evaluation budget, does \textsc{AscendOptimizer} outperform strong search baselines? (2) What is the contribution of each system component? (3) Can Optimization Rewind recover optimization knowledge that is scarce for the model---that is, structural rewrites the model rarely discovers or applies from bottleneck descriptions alone? (4) How do kernel refinement and tiling search contribute to optimization performance?

Appendix~\ref{app:exp_details} gives additional benchmark details and the full operator list.

\subsection{Experimental Setup}
\label{sec:hardware_metrics}

\noindent\textbf{Setup and correctness.}
Experiments run on Ascend 910B2 NPUs with CANN 8.3. All experiments that use an LLM use DeepSeek-V3.2, including rewind candidate generation, kernel rewriting, and tiling mutation in \textsc{AscendOptimizer}. The embedding model used for experience retrieval is \texttt{text-embedding-3-small}. All methods use the same toolchain, stream settings, synchronization, and correctness harness. We compare NPU outputs with CPU references using absolute/relative tolerance checks following official CANN example policies.

\noindent\textbf{Benchmark and stratification.}
We construct the benchmark from the official AscendC \texttt{cann-ops}\footnote{\url{https://gitee.com/ascend/cann-ops}} repository and use its implementations as baselines. We retain operators that compile, execute, pass CPU-reference checks, and show stable baseline latency under repeated profiling, yielding 101 operators for all reported experiments. We then stratify these operators into three levels by \emph{engineering complexity}, based on the computation path, input-output dependencies, data orchestration, and parallel scheduling rather than functional category or measured speed: Level~1 has 25 elementary math, logical comparison, and simple tensor-construction operators with regular memory access, such as \texttt{add\_custom}, \texttt{sqrt}, and \texttt{logical\_or}; Level~2 has 63 \texttt{foreach}-family, activation, normalization, fused, pooling, and upsampling operators, such as \texttt{foreach\_pow\_scalar\_and\_tensor}, \texttt{gelu}, \texttt{avg\_pool3\_d}, and \texttt{add\_layer\_norm\_grad}; and Level~3 has 13 operators with complex parallelization, irregular memory access, elaborate data orchestration, or heavy reliance on low-level optimization primitives, such as \texttt{flash\_attention\_score\_with\_large\_head\_dim}, \texttt{matmul\_all\_reduce}, \texttt{bev\_pool}, and \texttt{scatter\_reduce}. Main results are reported by this engineering-complexity level; and full operator list are detailed in Appendix~\ref{app:exp_details}.

\noindent\textbf{Metrics and budget.}
We report latency, geometric-mean speedup over \texttt{cann-ops},
\begin{equation}
\mathrm{speedup}(op) = \frac{T_{\mathrm{baseline}}(op)}{T_{\mathrm{gen}}(op)}.
\end{equation}
and $\mathrm{fast}_p$, the fraction of operators with speedup greater than $p$:
\begin{equation}
\mathrm{fast}_p = \frac{\left|\left\{ op \mid \mathrm{speedup}(op) > p \right\}\right|}{\left|\mathcal{O}\right|}.
\end{equation}
Each hardware evaluation is one compile--correctness--profiling attempt; failures and timeouts count toward the budget but do not update the incumbent. In our environment, each such hardware evaluation takes roughly two minutes on average, so the evaluation budget also reflects a practical wall-clock optimization cost. All main-comparison methods receive 230 evaluations per operator. For \textsc{AscendOptimizer}, 20 are attributed to rewind-stage experience construction and 210 to online optimization.

\subsection{Overall Performance}

Table~\ref{tab:main_performance} answers the first question by comparing \textsc{AscendOptimizer} with Best-of-$N$ sampling (BoN) and OpenEvolve~\cite{openevolve} under the same 230-evaluation budget per operator. For each engineering-complexity level defined in Section~\ref{sec:hardware_metrics}, we report geometric-mean speedup (GM) and the $\text{fast}_x$ ratios for $x \in \{1.2, 1.4, 1.8, 2.0\}$, where higher is better.
Across all three levels, \textbf{AscendOptimizer} achieves the best GM and the best or tied-best tail-speedup ratios. The advantage is largest on level3, where GM reaches 1.89 versus 1.38 for BoN and 1.45 for OpenEvolve; \SI{53.85}{\percent} of these hardest operators exceed $1.2\times$ and \SI{30.77}{\percent} exceed $2.0\times$. This trend indicates that experience-guided kernel rewriting and tiling/resource search are most valuable when operators have richer control flow, memory orchestration, and scheduling constraints.

\begin{table}[htbp]
\centering
\caption{\textbf{Main performance.} GM is geometric-mean speedup over \texttt{cann-ops}; fast$_p$ is the fraction of operators exceeding $p\times$ speedup.}
\label{tab:main_performance}
\setlength{\tabcolsep}{3pt}
\small
\begin{tabularx}{\columnwidth}{@{}l c l *{5}{>{\centering\arraybackslash}X}@{}}
\toprule
\textbf{Level} & \textbf{Tasks} & \textbf{Method} & \textbf{GM} $\uparrow$ & \textbf{fast$_{1.2}$} $\uparrow$ & \textbf{fast$_{1.4}$} $\uparrow$ & \textbf{fast$_{1.8}$} $\uparrow$ & \textbf{fast$_{2.0}$} $\uparrow$ \\
\midrule

\multirow{3}{*}{level1}
& \multirow{3}{*}{25} & BoN & \heatcell{1.04}{1.04}{1.24}{1.04} & \heatcell{8.00}{8.00}{12.00}{8.00\%} & \heatcell{4.00}{4.00}{12.00}{4.00\%} & \heatcell{4.00}{4.00}{12.00}{4.00\%} & \heatcell{0.00}{0.00}{8.00}{0.00\%} \\
& & OpenEvolve & \heatcell{1.11}{1.04}{1.24}{1.11} & \heatcell{8.00}{8.00}{12.00}{8.00\%} & \heatcell{8.00}{4.00}{12.00}{8.00\%} & \heatcell{4.00}{4.00}{12.00}{4.00\%} & \heatcell{4.00}{0.00}{8.00}{4.00\%} \\
& & \textbf{AscendOptimizer} & \heatcell{1.24}{1.04}{1.24}{\textbf{1.24}} & \heatcell{12.00}{8.00}{12.00}{\textbf{12.00\%}} & \heatcell{12.00}{4.00}{12.00}{\textbf{12.00\%}} & \heatcell{12.00}{4.00}{12.00}{\textbf{12.00\%}} & \heatcell{8.00}{0.00}{8.00}{\textbf{8.00\%}} \\

\midrule

\multirow{3}{*}{level2}
& \multirow{3}{*}{63} & BoN & \heatcell{1.05}{1.03}{1.10}{1.05} & \heatcell{6.35}{4.76}{14.29}{6.35\%} & \heatcell{3.17}{0.00}{6.35}{3.17\%} & \heatcell{3.17}{0.00}{6.35}{3.17\%} & \heatcell{1.59}{0.00}{6.35}{1.59\%} \\
& & OpenEvolve & \heatcell{1.03}{1.03}{1.10}{1.03} & \heatcell{4.76}{4.76}{14.29}{4.76\%} & \heatcell{0.00}{0.00}{6.35}{0.00\%} & \heatcell{0.00}{0.00}{6.35}{0.00\%} & \heatcell{0.00}{0.00}{6.35}{0.00\%} \\
& & \textbf{AscendOptimizer} & \heatcell{1.10}{1.03}{1.10}{\textbf{1.10}} & \heatcell{14.29}{4.76}{14.29}{\textbf{14.29\%}} & \heatcell{6.35}{0.00}{6.35}{\textbf{6.35\%}} & \heatcell{6.35}{0.00}{6.35}{\textbf{6.35\%}} & \heatcell{6.35}{0.00}{6.35}{\textbf{6.35\%}} \\

\midrule

\multirow{3}{*}{level3}
& \multirow{3}{*}{13} & BoN & \heatcell{1.38}{1.38}{1.89}{1.38} & \heatcell{46.15}{38.46}{53.85}{46.15\%} & \heatcell{38.46}{30.77}{46.15}{38.46\%} & \heatcell{23.08}{23.08}{30.77}{23.08\%} & \heatcell{15.38}{15.38}{30.77}{15.38\%} \\
& & OpenEvolve & \heatcell{1.45}{1.38}{1.89}{1.45} & \heatcell{38.46}{38.46}{53.85}{38.46\%} & \heatcell{30.77}{30.77}{46.15}{30.77\%} & \heatcell{23.08}{23.08}{30.77}{23.08\%} & \heatcell{23.08}{15.38}{30.77}{23.08\%} \\
& & \textbf{AscendOptimizer} & \heatcell{1.89}{1.38}{1.89}{\textbf{1.89}} & \heatcell{53.85}{38.46}{53.85}{\textbf{53.85\%}} & \heatcell{46.15}{30.77}{46.15}{\textbf{46.15\%}} & \heatcell{30.77}{23.08}{30.77}{\textbf{30.77\%}} & \heatcell{30.77}{15.38}{30.77}{\textbf{30.77\%}} \\

\bottomrule
\end{tabularx}
\end{table}

\begin{figure}[t]
\centering
\includegraphics[width=0.5\columnwidth]{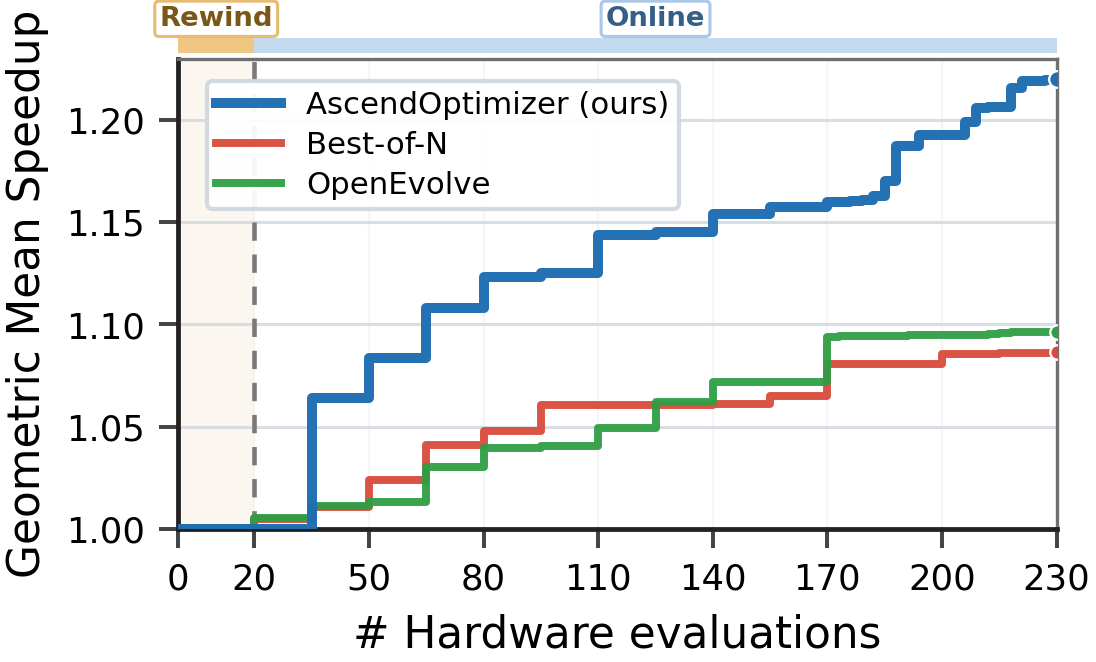}
\caption{\textbf{Optimization progress under hardware-evaluation budget.} AscendOptimizer remains ahead of Best-of-$N$ sampling and OpenEvolve across the evaluated budgets.}
\label{fig:budget_progress}
\end{figure}

Figure~\ref{fig:budget_progress} provides supporting evidence by plotting geometric-mean speedup as the hardware-evaluation budget increases. We use it as a consistency check for the main comparison rather than as a standalone conclusion about budget use: within the evaluated budget range, \textsc{AscendOptimizer} remains ahead of BoN and OpenEvolve. Overall, these results show that \textsc{AscendOptimizer} improves over both baselines under the same evaluation budget, with the clearest advantage on higher-complexity operators.

\subsection{Component Analysis}

Table~\ref{tab:ablation_performance} answers the second question by comparing independently run variants under the same 230-evaluation budget per operator; the rows are not sequential stages that inherit one another's intermediate results. Stage~\Rmnum{1} is the strongest single component, reaching 1.16 GM and showing that retrieval-guided kernel refinement is the primary source of structural improvement. Removing Experience lowers GM from 1.16 to 1.09 and reduces fast$_{2.0}$ from 7.92\% to 3.88\%, which shows that retrieved rewind experience is not merely auxiliary prompt context. Stage~\Rmnum{2} alone reaches only 1.02 GM, suggesting that host-side tiling search has limited room when the original kernel structure remains unchanged. The full system reaches 1.21 GM and increases fast$_{2.0}$ to 9.90\%, indicating that tiling search is most useful after Stage~\Rmnum{1} has improved the kernel. Thus, kernel refinement provides the dominant gain, retrieval is important within Stage~\Rmnum{1}, and tiling search adds further benefit on top of the refined kernel.

\begin{table}[!t]
\centering

\caption{\textbf{Component ablation.} Results are computed over all 101 operators. Each variant is run independently with the same 230-evaluation budget per operator. Columns $>p$ report fast$_p$, the fraction of operators with speedup greater than $p\times$. Higher is better.}
\label{tab:ablation_performance}
\setlength{\tabcolsep}{3.5pt}
\renewcommand{\arraystretch}{1.08}
\small
\begin{tabularx}{\columnwidth}{@{}l c c c *{5}{>{\centering\arraybackslash}X}@{}}
\toprule
\textbf{Variant} & \multicolumn{3}{c}{\textbf{Enabled components}} & \multicolumn{5}{c}{\textbf{Performance}} \\
\cmidrule(lr){2-4}\cmidrule(l){5-9}
& \textbf{Rewrite} & \textbf{Experience} & \textbf{Tiling} & \textbf{GM} $\uparrow$ & \textbf{$>1.2$} $\uparrow$ & \textbf{$>1.4$} $\uparrow$ & \textbf{$>1.8$} $\uparrow$ & \textbf{$>2.0$} $\uparrow$ \\
\midrule
Stage~\Rmnum{1} only & \cmark & \cmark & \xmark & 1.16 & 18.81\% & 12.87\% & 9.90\% & 7.92\% \\
Stage~\Rmnum{1} w/o Experience & \cmark & \xmark & \xmark & 1.09 & 12.62\% & 8.74\% & 5.83\% & 3.88\% \\
Stage~\Rmnum{2} only & \xmark & \xmark & \cmark & 1.02 & 2.97\% & 1.98\% & 0.00\% & 0.00\% \\
\textbf{AscendOptimizer} & \cmark & \cmark & \cmark & \textbf{1.21} & \textbf{18.81\%} & \textbf{12.87\%} & \textbf{10.89\%} & \textbf{9.90\%} \\
\bottomrule
\end{tabularx}
\end{table}

\subsection{Rewind Mines Model-Scarce Optimization Knowledge}
\label{sec:case_upsample_refactor}

\begin{figure}[!htbp]
\centering
\includegraphics[width=0.62\linewidth]{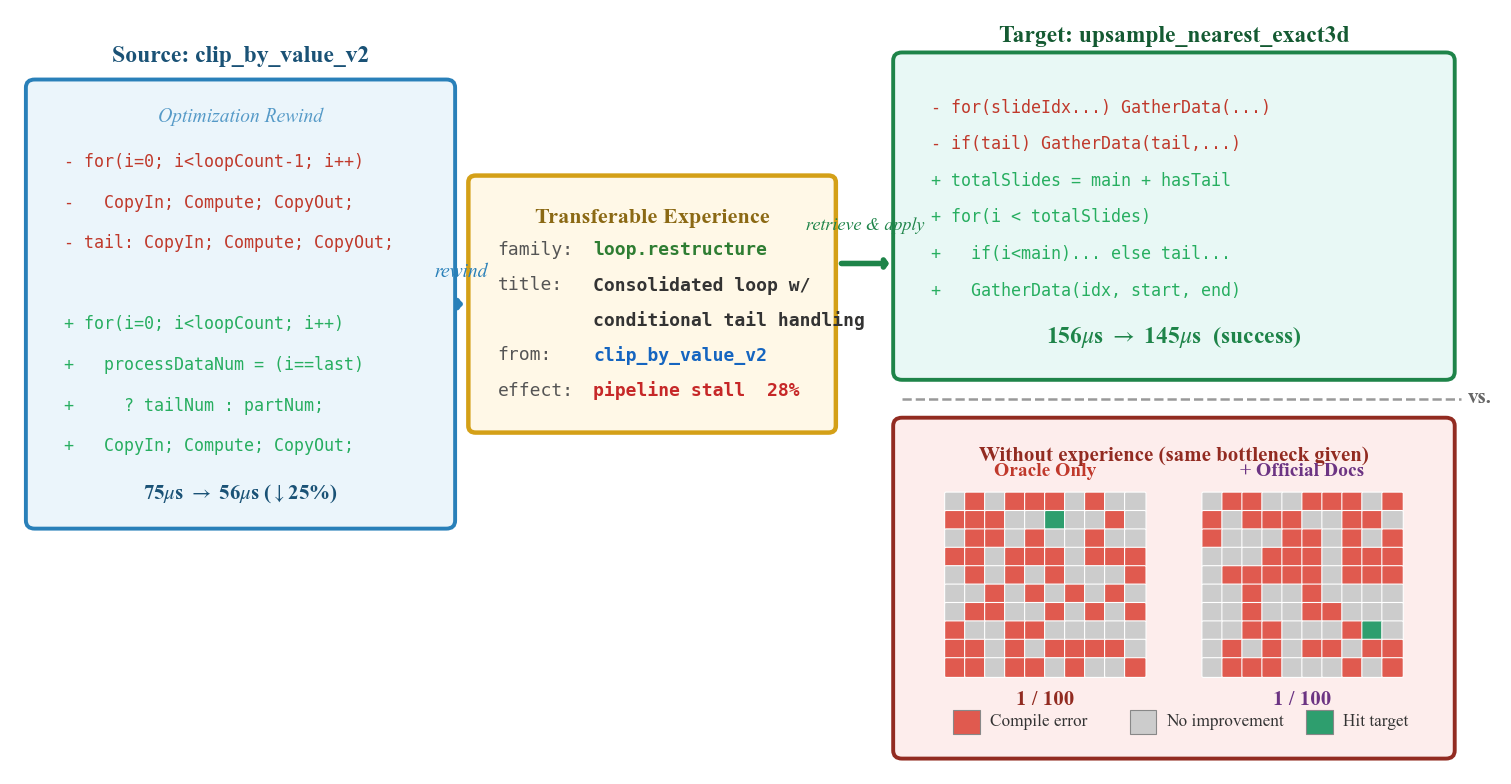}
\caption{Cross-operator transfer for \texttt{upsample\_nearest\_exact3d}. A loop-consolidation pattern mined from \texttt{clip\_by\_value\_v2} transfers to the target operator; without retrieval, oracle bottleneck guidance rarely finds the same rewrite.}
\label{fig:case_study_bridge}
\end{figure}

We next answer the third question with a concrete cross-operator example showing that Optimization Rewind extracts knowledge the model rarely produces from bottleneck text alone (Figure~\ref{fig:case_study_bridge}).

\noindent\textbf{Experience mining from a source operator.}
During rewind exploration of \texttt{clip\_by\_value\_v2}, AscendOptimizer identifies a loop-consolidation optimization by observing what happens when it is removed. The high-performance implementation folds separate tail processing into the main loop: \texttt{for(i<loopCount-1)\{...\}; tail: CopyIn; Compute; CopyOut;} becomes a unified \texttt{for(i<loopCount)} with \texttt{(i==last) ? tailNum : partNum}. Hardware validation confirms that this motif matters, reducing pipeline stalls by 28.6\% and latency from \SI{75}{\micro\second} to \SI{56}{\micro\second}, a 25\% improvement. The resulting pattern, titled \emph{Consolidated loop structure with conditional tail handling}, is stored with its causal mechanism, applicability conditions, and code-level rewrite template.

\noindent\textbf{Cross-operator retrieval and application.} For a structurally unrelated operator, \texttt{upsample\_nearest\_exact3d}, whose initial latency is \SI{156}{\micro\second}, profiling localizes the bottleneck to the \texttt{GatherData} hot path, where separate main-slide and tail-slide handling causes vector-pipeline stalls and low vector utilization. The diagnosis identifies where the problem is, but not the structural rewrite. Using it as the retrieval query, AscendOptimizer retrieves the pattern mined from \texttt{clip\_by\_value\_v2} and applies the analogous transformation: a separate \texttt{for(slideIdx)} loop plus post-loop \texttt{if(tail) GatherData(tail,...)} becomes one \texttt{for(i<totalSlides)} loop with internal conditional dispatch. This produces the first major performance drop, from \SI{156}{\micro\second} to \SI{145}{\micro\second}.

\noindent\textbf{Counterfactual: the knowledge is model-scarce.}
To isolate the effect of retrieved experience, we give the optimizer the same oracle bottleneck description, identifying the \texttt{GatherData} pipeline stall, but no retrieved experience. Across 100 independent attempts, only 1 reaches the \SI{145}{\micro\second} target. A stronger baseline that also receives the \href{https://www.hiascend.com/document/detail/zh/CANNCommunityEdition/83RC1alpha002/opdevg/ascendcbestP/atlas_ascendc_best_practices_10_00010.html}{official Ascend~C Best Practices documentation} still has a 1/100 hit rate. Without retrieval, the optimizer usually proposes broader but ineffective edits such as \emph{Vector Pipeline Barrier Reduction} or generic scalar cleanup, and only rarely reaches the specific loop-restructuring transformation.

The key point is not that the full system discovers a different bottleneck. The no-experience ablation also identifies related issues such as scalarized gather or offset processing and poor vector utilization. What rewind-derived retrieval adds is the missing link from diagnosis to an \emph{actionable structural transformation}: the exact loop-consolidation rewrite that unguided LLM generation and static vendor documentation rarely produce. In this sense, Optimization Rewind turns latent structure in existing kernels into explicit, retrievable knowledge that is scarce for the model during direct optimization.

\subsection{Different Roles of Kernel and Tiling Optimization}
\label{sec:case_matmul_all_reduce}
\begin{figure}[H]
\centering
\vspace{-0.4em}
\begin{subfigure}[t]{0.96\linewidth}
  \centering
  \includegraphics[width=0.8\linewidth]{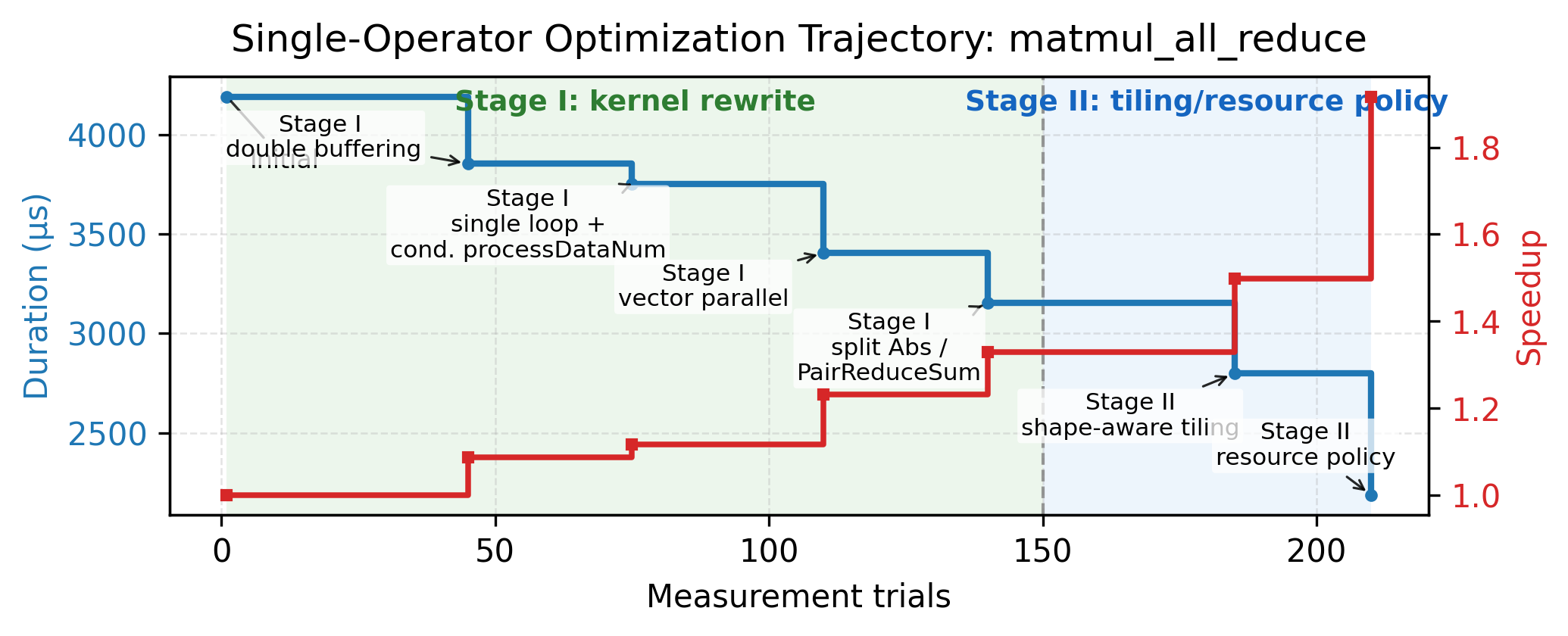}
  \caption{Trajectory for \texttt{matmul\_all\_reduce}: Stage~\Rmnum{1} improves kernel structure; Stage~\Rmnum{2} improves tiling and resource policy.}
\end{subfigure}

\vspace{-0.3em}

\begin{subfigure}[t]{0.96\linewidth}
\newcommand{\drem}[1]{{\color{red!70!black}\texttt{#1}}}%
\newcommand{\dadd}[1]{{\color{green!40!black}\texttt{#1}}}%
\begin{minipage}[t]{0.315\linewidth}
\begin{tcolorbox}[
  enhanced,colback=white,colframe=green!50!black,
  boxrule=0.4pt,arc=1.5pt,
  fonttitle=\bfseries\scriptsize,
  title={\raisebox{0pt}[8pt][2pt]{\Rmnum{1}:~1.23$\times$$\to$1.33$\times$~~\upshape split loops}},
  coltitle=white,colbacktitle=green!45!black,
  left=1pt,right=1pt,top=1pt,bottom=1pt,
  before skip=0pt,after skip=0pt
]
\fontsize{5.3pt}{6.2pt}\selectfont\raggedright
\drem{-~for~(i=0;~i<blockCnt;~i++)~\{}\\
\drem{-~~curM=isLastM?mTail:baseM;}\\
\drem{-~~curN=isLastN?nTail:baseN;}\\
\drem{-~~mm.SetSingleShape(curM,curN,K);}\\
\drem{-~~mm.Iterate();~\}}\\[0.5pt]
\dadd{+~while~(full\_block)~\{}\\
\dadd{+~~mm.SetSingleShape(baseM,baseN,K);}\\
\dadd{+~~mm.Iterate();~processed++;~\}}\\
\dadd{+~while~(M\_tail)~\{}\\
\dadd{+~~mm.SetSingleShape(mTail,baseN,K);}\\
\dadd{+~~mm.Iterate();~processed++;~\}}\\
\dadd{+~while~(N\_tail)~\{...\}}\\
\dadd{+~while~(MN\_tail)~\{...\}}
\end{tcolorbox}
\end{minipage}
\hfill
\begin{minipage}[t]{0.315\linewidth}
\begin{tcolorbox}[
  enhanced,colback=white,colframe=blue!50!black,
  boxrule=0.4pt,arc=1.5pt,
  fonttitle=\bfseries\scriptsize,
  title={\raisebox{0pt}[8pt][2pt]{\Rmnum{2}:~1.33$\times$$\to$1.50$\times$~~\upshape shape-aware tiling}},
  coltitle=white,colbacktitle=blue!45!black,
  left=1pt,right=1pt,top=1pt,bottom=1pt,
  before skip=0pt,after skip=0pt
]
\fontsize{5.3pt}{6.2pt}\selectfont\raggedright
\drem{-~tileM=ComputeTileM(m,cores,}\\
\drem{-~~~~defaultTileM,128);}\\
\drem{-~//~M-only~size~thresholds}\\[0.5pt]
\dadd{+~tileM=ComputeTileM(m,n,k,}\\
\dadd{+~~~~cores,isTransA,isTransB,}\\
\dadd{+~~~~dtypeSize);}\\
\dadd{+~intensity=2*m*n*k/totalBytes;}\\
\dadd{+~if~(memory\_bound)}\\
\dadd{+~~baseTile=small\_reuse(m);}\\
\dadd{+~else~if~(compute\_bound)}\\
\dadd{+~~baseTile=large\_tile(m);}\\
\dadd{+~comm=m*n*dtypeSize/cores;}\\
\dadd{+~baseTile=adjust\_allreduce(comm);}
\end{tcolorbox}
\end{minipage}
\hfill
\begin{minipage}[t]{0.315\linewidth}
\begin{tcolorbox}[
  enhanced,colback=white,colframe=blue!50!black,
  boxrule=0.4pt,arc=1.5pt,
  fonttitle=\bfseries\scriptsize,
  title={\raisebox{0pt}[8pt][2pt]{\Rmnum{2}:~1.50$\times$$\to$1.92$\times$~~\upshape resource policy}},
  coltitle=white,colbacktitle=blue!45!black,
  left=1pt,right=1pt,top=1pt,bottom=1pt,
  before skip=0pt,after skip=0pt
]
\fontsize{5.3pt}{6.2pt}\selectfont\raggedright
\drem{-~blockDim=min(tiles,cores);}\\
\drem{-~if~(matSize<10000)}\\
\drem{-~~blockDim=min(blockDim,4);}\\
\drem{-~//~fixed~thresholds~only}\\[0.5pt]
\dadd{+~for~(d=1;~d<=maxCores;~d++)~\{}\\
\dadd{+~~bal=1-extra/tiles;}\\
\dadd{+~~util=tiles/d;}\\
\dadd{+~~comm=1/(1+.05*log2(d+1));}\\
\dadd{+~~score=bal*util*comm*cache;}\\
\dadd{+~~if~(score>best)~bestDim=d;~\}}\\
\dadd{+~blockDim=bestDim;}
\end{tcolorbox}
\end{minipage}
\caption{Representative diffs: Stage~\Rmnum{1} specializes kernel loops; Stage~\Rmnum{2} replaces fixed host-side heuristics with shape-aware tiling and scored resource allocation.}
\end{subfigure}
\caption{Code-and-curve case study for \texttt{matmul\_all\_reduce}, showing the different roles of kernel refinement and host-side tiling search.}
\label{fig:matmul_all_reduce_stage_case}
\end{figure}

Finally, we examine \texttt{matmul\_all\_reduce} to answer what kernel refinement and tiling search do in the optimized system. The first outer-loop Stage~\Rmnum{1} pass reduces the score from \num{4192} to \num{3154} by rewriting the kernel execution path itself. The main edits are structural: the kernel separates full-block work from tail cases, restructures reduction-style loops, adjusts pipeline depth and buffering decisions, and exposes more vector-core parallelism with less synchronization overhead. In other words, Stage~\Rmnum{1} changes how the hot path runs inside the kernel. Starting from that Stage~\Rmnum{1}-optimized kernel, the subsequent Stage~\Rmnum{2} pass further lowers the refreshed score from \num{3144} to \num{2187} by changing the host-side policies that feed the kernel. At this stage, the focus is no longer kernel control flow but launch and resource selection: the tiler chooses \texttt{tileM} using matrix shape, compute intensity, transpose flags, and all-reduce communication cost; chooses \texttt{blockDim} under load-balance and communication penalties; sizes UB/L1 buffers and workspace hierarchically; and emits pipeline, prefetch, and double-buffer hints. \Cref{fig:matmul_all_reduce_stage_case} connects this trajectory to representative diff hunks and shows the separation clearly: kernel refinement changes the internal execution structure, while tiling search tunes the surrounding tiling and resource policy for that refined kernel.

\FloatBarrier

\vspace{-2em}
\section{Conclusion}
\label{sec:conclusion}
This paper studies training-free optimization of AscendC operators under severe scarcity of expert knowledge and paired training data. AscendOptimizer bootstraps a shared experience bank through controlled Optimization Rewind, uses it for retrieval-guided kernel refinement, and then performs hardware-in-the-loop tiling search under the refined kernel. On 101 operators, this design consistently outperforms strong search and agent baselines; limitations and broader impacts are in Appendix~\ref{app:limitations_impact}.

\bibliography{example_paper}
\bibliographystyle{plain}

\newpage
\appendix
\onecolumn

\section{Additional Experimental Details}
\label{app:exp_details}

\noindent\textbf{Final benchmark size.}
After benchmark filtering and shape adjustments, we retain 101 operators for all reported experiments.

\noindent\textbf{Benchmark construction and filtering.}
We start from the \texttt{cann-ops} AscendC repository and use the provided implementations as baselines. During preparation, we verify compilability, check numerical correctness against CPU references, and remove operators that fail compilation, execution, correctness, or baseline-latency stability checks.

\noindent\textbf{Input-shape adjustment.}
For very small workloads, hardware timing can be noisy, so we moderately increase input shapes for a subset of operators. Shape changes preserve operator semantics; the baseline and all optimized variants use the same adjusted shape; and the scaling factor is kept small.

\noindent\textbf{Evaluation paradigm and data overlap.}
The offline rewind phase is benchmark-wide: each operator contributes 20 rewind evaluations before the validated trajectories are aggregated into the shared bank. No model weights are updated. This follows a transductive system-autotuning objective: finding low latency for a given workload on a fixed hardware/software stack, rather than claiming zero-shot predictive generalization.

\noindent\textbf{Operator list by level.}
The following table gives the full operator assignment under the engineering-complexity grading defined in Section~\ref{sec:hardware_metrics}.

\begin{tabular}{@{}l p{0.78\linewidth}@{}}
\hline
\textbf{Level} & \textbf{Operator} \\
\hline
Level~1 (25) & add\_custom, angle\_v2, arange, cast, cos, equal, eye, fill, greater\_equal, heaviside, icamax, icamin, is\_inf, isamin, less, less\_equal, logical\_not, logical\_or, muls, neg, reciprocal, reshape, rsqrt, sqrt, trunc \\
Level~2 (63) & foreach\_acos, foreach\_add\_list, foreach\_add\_scalar, foreach\_add\_scalar\_list, foreach\_addcdiv\_scalar, foreach\_addcmul\_scalar, foreach\_addcmul\_scalar\_list, foreach\_asin, foreach\_copy, foreach\_div\_list, foreach\_div\_scalar, foreach\_erf, foreach\_erfc, foreach\_exp, foreach\_lerp\_scalar, foreach\_log2, foreach\_maximum\_scalar, foreach\_minimum\_scalar, foreach\_mul\_list, foreach\_mul\_scalar, foreach\_neg, foreach\_pow\_scalar, foreach\_pow\_scalar\_and\_tensor, foreach\_reciprocal, foreach\_round\_off\_number, foreach\_sigmoid, foreach\_sign, foreach\_sinh, foreach\_sqrt, foreach\_sub\_scalar, foreach\_tan, foreach\_tanh, foreach\_zero\_inplace, ge\_glu\_grad\_v2, ge\_glu\_v2, ge\_glu\_v3, gelu\_quant, swi\_glu, swi\_glu\_grad, pad\_v3\_grad\_replication, reflection\_pad3d\_grad, strided\_slice\_assign\_v2, upsample\_bilinear2d\_grad, upsample\_nearest\_exact3d, addcdiv, addcmul, clip\_by\_value\_v2, fast\_gelu\_grad, gcd, gelu, gelu\_grad, lerp, lin\_space, mul\_sigmoid, swish, tril, triu, adaptive\_avg\_pool3d\_grad, avg\_pool3\_d, add\_layer\_norm\_grad, add\_rms\_norm\_dynamic\_quant, add\_rms\_norm\_quant, group\_norm\_swish \\
Level~3 (13) & motion\_compensation, upsample\_bilinear2d\_aa, upsample\_bilinear2d\_aa\_backward, upsample\_trilinear3d\_backward, add\_sigmoid\_mul\_reduce\_sum\_d, bev\_pool, flash\_attention\_score\_with\_large\_head\_dim, matmul\_all\_reduce, matmul\_api\_constant, scatter\_reduce, strideslice\_neg\_concat\_v2, complex\_mat\_mul, foreach\_non\_finite\_check\_and\_unscale \\
\hline
\end{tabular}

\section{Method Details and Experience Bank Analysis}
\label{app:method_experience}
This section provides additional analysis of the optimization experience bank together with detailed method snippets.

\subsection{Rewind as Local One-Factor-at-a-Time Screening}
\label{app:rewind_formalization}

Following the problem formulation in Section~\ref{sec:method}, an operator is
represented as $\mathcal{O}=\langle\mathcal{T},\mathcal{K},\mathcal{S}\rangle$.
During the offline rewind phase, we fix the host-side tiling function
$\mathcal{T}$ and focus on the kernel-side implementation space.

Let $\mathbb{K}_{\mathcal{S}}$ denote the set of kernel implementations that
realize the operator semantics under static attributes $\mathcal{S}$ and pass
correctness validation under the current execution harness. For a fixed tiling
function $\mathcal{T}$, we define the hardware utility of a kernel as
\begin{equation}
    U_{\mathcal{T}}(\mathcal{K}) = -\mathcal{L}(\mathcal{K}\mid\mathcal{T}),
    \label{eq:rewind_utility}
\end{equation}
where $\mathcal{L}(\mathcal{K}\mid\mathcal{T})$ is the measured latency
introduced in Eq.~(\ref{eq:problem_objective}); higher utility corresponds to
lower latency.

We view a kernel as activating a set of recognizable optimization motifs
$A\subseteq\Omega$, where $\Omega$ is a finite catalog of named motifs such as
double buffering, vectorized data movement, loop-tail consolidation,
synchronization elimination, or redundant-write removal. Equivalently, $A$ is
represented by a binary motif vector $z\in\{0,1\}^{|\Omega|}$ with $z_m=1$ iff
motif $m$ is active. Multiple kernels in $\mathbb{K}_{\mathcal{S}}$ may share
the same active set $A$; we therefore write
$\mathcal{K}_{A}\in\mathbb{K}_{\mathcal{S}}$ for any representative
implementation whose active motif set equals $A$, with the choice of
representative determined by the inverse agent and the validation harness.

A rewind action for motif $m$ is a \emph{partial} map
\begin{equation}
    r_m : \mathbb{K}_{\mathcal{S}} \rightharpoonup \mathbb{K}_{\mathcal{S}},
    \label{eq:rewind_operator}
\end{equation}
left undefined (written $r_m(\mathcal{K})=\bot$) whenever the proposed
de-optimized candidate fails compilation, fails correctness validation, or
cannot be isolated as a semantic-preserving rewrite of motif $m$. When
$\mathcal{K}=\mathcal{K}_{A}$ with $m\in A$, a successful rewind yields a
neighboring representative
\begin{equation}
    r_m(\mathcal{K}_{A}) \;=\; \mathcal{K}_{A\setminus\{m\}}.
    \label{eq:motif_removal}
\end{equation}

The measured effect of this single-motif removal is
\begin{equation}
    \widehat{\Delta}_m(\mathcal{K}_{A};\mathcal{T})
    \;=\;
    U_{\mathcal{T}}(\mathcal{K}_{A}) - U_{\mathcal{T}}(\mathcal{K}_{A\setminus\{m\}})
    \;=\;
    \mathcal{L}(\mathcal{K}_{A\setminus\{m\}}\mid\mathcal{T}) - \mathcal{L}(\mathcal{K}_{A}\mid\mathcal{T}),
    \label{eq:local_delta}
\end{equation}
so $\widehat{\Delta}_m>0$ indicates that motif $m$ contributes positively to
performance at the basepoint $\mathcal{K}_{A}$. Equation~(\ref{eq:local_delta})
is a local one-factor-at-a-time (OAT) finite difference on the discrete motif
space, evaluated at a single strong-implementation basepoint per operator.
This is in the spirit of elementary-effect screening in sensitivity
analysis, with two practical differences: (i) the perturbation flips a binary
motif variable rather than shifting a continuous input, and (ii) we use the
local difference as a \emph{validated screening signal}, not as an estimator
of a global elementary-effect distribution.

We retain a rewind-derived experience only when
\begin{equation}
    \widehat{\Delta}_m(\mathcal{K}_{A};\mathcal{T}) > \tau_{\mathrm{noise}},
    \label{eq:accept_delta}
\end{equation}
where $\tau_{\mathrm{noise}}$ is calibrated above profiling noise on the target
hardware. Candidates with $r_m(\mathcal{K}_{A})=\bot$ or with effects below
$\tau_{\mathrm{noise}}$ are discarded. Each accepted pair
$(\mathcal{K}_{A},\mathcal{K}_{A\setminus\{m\}})$, together with its validated
effect $\widehat{\Delta}_m$, is the raw material from which the structured
experience tuple $\mathcal{M}$ in Eq.~(\ref{eq:experience_tuple}) is distilled.
Thus the experience bank stores hardware-validated local screening signals
rather than assumed optimization rules.

\subsection{Optimization Algorithms}

\begin{algorithm}[H]
\small
\caption{Shared Experience Bank Construction via Optimization Rewind}
\label{alg:rewind_bank}
\begin{algorithmic}[1]
\Require Benchmark operator set $\mathcal{O}$, rewind budget $B_{r}$
\Output Shared experience bank $\mathcal{B}$
\State $\mathcal{B} \gets \emptyset$
\For{each operator $o \in \mathcal{O}$}
    \State $\mathcal{K}_{\text{curr}} \gets$ strongest available implementation of $o$
    \For{$b = 1$ to $B_{r}$}
        \State propose semantically meaningful rewind candidate $\tilde{\mathcal{K}}$
        \If{$\tilde{\mathcal{K}}$ compiles, passes correctness, and is slower than $\mathcal{K}_{\text{curr}}$}
            \State distill tuple $\mathcal{M}$ from $(\mathcal{K}_{\text{curr}}, \tilde{\mathcal{K}})$ and profiling deltas
            \State $\mathcal{B} \gets \mathcal{B} \cup \{\mathcal{M}\}$; $\mathcal{K}_{\text{curr}} \gets \tilde{\mathcal{K}}$
        \EndIf
    \EndFor
\EndFor
\State \Return $\mathcal{B}$
\end{algorithmic}
\end{algorithm}

\begin{algorithm}[H]
\small
\caption{Per-Operator Online Optimization with Kernel Refinement and Tiling Search}
\label{alg:online_opt}
\begin{algorithmic}[1]
\Require Target operator $(\mathcal{T}^{(0)}, \mathcal{K}^{(0)}, \mathcal{S})$, shared bank $\mathcal{B}$, kernel steps $S$, tiling steps $U$
\Output Optimized pair $(\mathcal{T}^{\dagger}, \mathcal{K}^{\dagger})$
\State $\mathcal{T}_{\text{curr}} \gets \mathcal{T}^{(0)}$, $\mathcal{K}_{\text{curr}} \gets \mathcal{K}^{(0)}$
\For{$s = 1$ to $S$}
    \State $q \gets \textsc{DiagnoseBottleneck}(\mathcal{T}_{\text{curr}}, \mathcal{K}_{\text{curr}})$
    \State $\{\mathcal{M}_{i}\}_{i=1}^{k} \gets \textsc{RetrieveTopK}(\mathcal{B}, q)$
    \State $\tilde{\mathcal{K}} \gets \textsc{RewriteKernel}(\mathcal{K}_{\text{curr}}, \{\mathcal{M}_{i}\}_{i=1}^{k})$
    \If{$\tilde{\mathcal{K}}$ compiles, passes correctness, and improves $\mathcal{L}(\tilde{\mathcal{K}}\mid\mathcal{T}_{\text{curr}})$}
        \State $\mathcal{K}_{\text{curr}} \gets \tilde{\mathcal{K}}$
    \EndIf
\EndFor
\For{$u = 1$ to $U$}
    \State $\tilde{\mathcal{T}} \gets \textsc{MutateTiling}(\mathcal{T}_{\text{curr}}, \mathcal{S})$
    \If{$\tilde{\mathcal{T}}$ compiles, passes correctness, and improves $\mathcal{L}(\mathcal{K}_{\text{curr}}\mid\tilde{\mathcal{T}})$}
        \State $\mathcal{T}_{\text{curr}} \gets \tilde{\mathcal{T}}$
    \EndIf
\EndFor
\State \Return $(\mathcal{T}_{\text{curr}}, \mathcal{K}_{\text{curr}})$
\end{algorithmic}
\end{algorithm}

\lstdefinestyle{appendixcode}{
  basicstyle=\ttfamily\footnotesize,
  columns=fullflexible,
  keepspaces=true,
  showstringspaces=false,
  breaklines=true,
  breakatwhitespace=true,
  frame=single,
  framerule=0.3pt,
  rulecolor=\color{black!30},
  xleftmargin=0.5em,
  xrightmargin=0.5em,
  aboveskip=0.6em,
  belowskip=0.6em,
  numbers=left,
  numberstyle=\tiny\color{black!50},
  numbersep=6pt,
  tabsize=2,
  captionpos=b,
  language=C++,
  commentstyle=\color{green!35!black}\itshape,
  escapeinside={(*@}{@*)}
}

\subsection{Experience Bank Analysis}
Compared with official Ascend C best practices, the experience bank differs mainly in representation and scope. Official documentation is principle-level and written for human developers; the experience bank stores concrete, reusable optimization episodes tied to actual bottlenecks and code rewrites, so it can be retrieved and applied directly during automated optimization. In other words, the documentation says \emph{what is generally advisable}, while the experience bank records \emph{what specific transformation worked under what condition}.

Table~\ref{tab:experience_vs_docs} summarizes whether the bank stays within the scope of official guidance or extends beyond it. The bank certainly covers many motifs that are consistent with the documentation, such as tiling adjustment, DMA consolidation, double buffering, and vectorized reduction. However, it also goes beyond the documentation in several ways: it preserves operator-specific control-flow rewrites, scalar-index simplifications, early-exit or search-pruning tricks, and specialized SIMD paths that are useful in practice but are not usually stated as general vendor best practices.

\begin{table*}[t]
\centering
\caption{How the experience bank differs from and extends official Ascend C best practices}
\label{tab:experience_vs_docs}
\resizebox{\linewidth}{!}{%
\begin{tabular}{p{4.1cm} p{4.9cm} p{5.0cm}}
\toprule
\textbf{Aspect} & \textbf{Official Ascend C Best Practices} & \textbf{Experience Bank in This Work} \\
\midrule
Knowledge form
& Principle-level guidance for human developers
& Concrete optimization episodes with code-level rewrites and applicability conditions \\

Primary role
& Explain common optimization principles
& Support direct retrieval and transfer during automated operator optimization \\

Covered and aligned with docs
& Tiling, DMA efficiency, double buffering, vectorized reduction
& Recovers and operationalizes these documented principles in reusable optimization records \\

Beyond documented scope
& Usually does not enumerate long-tail operator-specific rewrites
& Includes loop restructuring, scalar index simplification, early-exit/search pruning, fine-grained synchronization tuning, and specialized SIMD paths \\

Granularity of reuse
& Broad recommendations
& Mechanism- and rewrite-level transferable patterns \\
\bottomrule
\end{tabular}%
}
\end{table*}
\FloatBarrier

\subsection{Method Details}\label{app:methodDet}
This subsection gives two implementation-level views of the optimizer. Figure~\ref{fig:appendix_tiling_code} shows how Stage~\Rmnum{2} turns an operator and its attributes into an evolvable host-side tiling template. The important parts are not the surrounding C++ syntax, but the shape-derived features, the block/core resource parameters, and the output fields that define the mutation interface. Figure~\ref{fig:appendix_optimization_thought} shows the complementary Stage~\Rmnum{1} representation: a rewind-validated experience record that stores the retrieval key, causal mechanism, hardware evidence, and transfer constraints for a concrete code rewrite.

\begin{tcolorbox}[
  enhanced,
  breakable,
  colback=white,
  colframe=black!30,
  boxrule=0.3pt,
  arc=1.2pt,
  left=0.5em,right=0.5em,top=0.5em,bottom=0.5em,
  title={Method details: base tiling function $\mathcal{T}_{\text{base}}$},
  fonttitle=\bfseries,
]
\lstset{
  style=appendixcode,
  emph={batchSize,dimSizeX,dimSizeSrc,strideSize,reduction,blockSize,blockNum,coreNum,blockDim,specialCase,params},
  emphstyle=\color{blue!70!black}\bfseries
}
\begin{lstlisting}
// (*@\colorbox{yellow!25}{\strut\textbf{FOCUS}}@*) evolvable region exposed to Stage II mutation/search
// # evolve_tiling_block_start
std::map<std::string, int64_t> AdaptiveTilingStrategy(const std::vector<int64_t>& xShape,
                                                      const std::vector<int64_t>& srcShape,
                                                      int64_t dim,
                                                      const std::string& reduceType,
                                                      bool includeSelf,
                                                      int64_t inputBytes)
{
    constexpr int64_t CORENUM = 1;
    constexpr int64_t BLOCK_BYTES_SIZE = 32;

    std::map<std::string, int64_t> params;
    // Guard clauses keep invalid candidates outside the search space.
    if (xShape.empty() || srcShape.empty()) {
        return params;
    }

    if (dim < 0 || dim >= static_cast<int64_t>(xShape.size()) ||
        dim >= static_cast<int64_t>(srcShape.size())) {
        return params;
    }

    // 1) Shape-derived features: fixed facts used by candidate tiling rules.
    int64_t batchSize = 1;
    for (int64_t i = 0; i < dim; ++i) {
        batchSize *= xShape[i];
    }

    const int64_t dimSizeX = xShape[dim];
    const int64_t dimSizeSrc = srcShape[dim];

    int64_t strideSize = 1;
    for (int64_t i = dim + 1; i < static_cast<int64_t>(xShape.size()); ++i) {
        strideSize *= xShape[i];
    }

    int64_t reduction = 0;
    if (reduceType == "sum") {
        reduction = 0;
    } else if (reduceType == "prod") {
        reduction = 1;
    } else if (reduceType == "mean") {
        reduction = 2;
    } else if (reduceType == "amax") {
        reduction = 3;
    } else if (reduceType == "amin") {
        reduction = 4;
    }

    // 2) Resource-granularity knobs: formulas/constants can be mutated.
    int64_t blockSize = (inputBytes > 0) ? (BLOCK_BYTES_SIZE / inputBytes) : BLOCK_BYTES_SIZE;
    if (blockSize <= 0) {
        blockSize = 1;
    }

    const int64_t blockNum = (strideSize + blockSize - 1) / blockSize;
    const int64_t coreNum = std::min<int64_t>(CORENUM, blockNum == 0 ? CORENUM : blockNum);

    const bool specialCase = (batchSize == 1 && dimSizeX == dimSizeSrc && !includeSelf &&
                              reduction == 4 && inputBytes == 4);
    const int64_t blockDim = specialCase ? coreNum : 0;

    // 3) Search interface: returned fields are consumed by host-side tiling.
    params["batchSize"] = batchSize;
    params["dimSizeX"] = dimSizeX;
    params["dimSizeSrc"] = dimSizeSrc;
    params["strideSize"] = strideSize;
    params["reduction"] = reduction;
    params["includeSelf"] = includeSelf ? 1 : 0;
    params["blockSize"] = blockSize;
    params["blockNum"] = blockNum;
    params["blockDim"] = blockDim;
    params["specialCaseFlag"] = specialCase ? 1 : 0;

    return params;
}
// # evolve_tiling_block_end

\end{lstlisting}
\tcblower
{\captionsetup{hypcap=false}
\captionof{figure}{Base tiling function $\mathcal{T}_{\text{base}}$ with evolution markers, synthesized from the operator code and attributes $\mathcal{S}$. The highlighted/comments indicate what readers should inspect: shape-derived facts, mutable block/core resource choices, and returned \texttt{params} fields that form the Stage~\Rmnum{2} search interface.}
\label{fig:appendix_tiling_code}}
\end{tcolorbox}

Figure~\ref{fig:appendix_tiling_code} is therefore a template boundary rather than a hand-written final tiling policy. The optimizer keeps the input validation and semantic features stable, then searches over the resource and dispatch choices exposed through \texttt{blockSize}, \texttt{blockNum}, \texttt{coreNum}, \texttt{blockDim}, and the returned parameter map. This makes the host-side search concrete enough to compile and profile, while still leaving the performance-sensitive policy choices open to mutation.

\begin{tcolorbox}[
  enhanced,
  breakable,
  colback=white,
  colframe=black!30,
  boxrule=0.3pt,
  arc=1.2pt,
  left=0.5em,right=0.5em,top=0.5em,bottom=0.5em,
  title={Method details: rewind-derived experience tuple},
  fonttitle=\bfseries,
]
\lstset{
  style=appendixcode,
  basicstyle=\ttfamily\scriptsize,
  emph={canonical_family,root_mechanism,causal_chain,profile_evidence,reusable_when,avoid_when,reuse_guidance,code_diff_excerpt},
  emphstyle=\color{blue!70!black}\bfseries
}
\begin{lstlisting}
{
  "op": "eye",
  "experience_type": "performance_recovery",
  "explore_mode": "degrade",

  // (*@\colorbox{yellow!25}{\strut\textbf{FOCUS 1}}@*) Retrieval key: groups this rewrite with related cases.
  "canonical_family": "kernel.memory.scatter",
  "optimization_title": "Eliminate redundant scattered zero writes",
  "mechanism_understood": true,

  // (*@\colorbox{yellow!25}{\strut\textbf{FOCUS 2}}@*) Mechanism: why the fast version is faster.
  "root_mechanism": "Output is already zero-initialized; only diagonal values need explicit writes.",
  "causal_chain": "remove scattered zero writes -> fewer L2 misses -> higher memory bandwidth -> shorter vector critical path",
  "slow_duration_us": 925024.5625,
  "fast_duration_us": 1193.783936,
  "improvement_direction": "degraded_is_slow",

  // (*@\colorbox{yellow!25}{\strut\textbf{FOCUS 3}}@*) Hardware evidence: the rewrite is validated by profiling.
  "profile_evidence": [
    "cache_l2.aiv_read_hit_rate_pct: 0.02 -> 87.5 (+437400.00%)",
    "critical_path.aiv_vector0_time_us_mean: 1193.17 -> 925006 (+77424.98%)",
    "latency.task_duration_us: 1193.78 -> 925025 (+77386.76%)"
  ],
  "affected_metric_groups": ["cache_l2", "critical_path", "latency"],

  // (*@\colorbox{yellow!25}{\strut\textbf{FOCUS 4}}@*) Transfer contract: when to reuse or avoid this rule.
  "retrieval_keywords": ["diagonal matrix", "scattered writes", "pre-initialized buffer"],
  "reusable_when": [
    "Eye-like diagonal initialization",
    "Output buffer is guaranteed to be zero-initialized"
  ],
  "avoid_when": [
    "Output buffer is not known to be zero-initialized"
  ],
  "reuse_guidance": [
    "Remove scattered SetValue calls that write non-diagonal zeros",
    "Keep only yGm.SetValue(index, 1) for diagonal positions"
  ],

  // Code evidence: rewind adds the slow behavior, so the inverse is the optimization.
  "code_diff_excerpt": [
    "  yGm.SetValue(index, 1);",
    "+ for (int32_t k = 0; k < num_columns; ++k) {",
    "+   if (k != i) {",
    "+     yGm.SetValue(i * num_columns + k, 0);",
    "+   }",
    "+ }  // redundant scattered zero writes"
  ]
}
\end{lstlisting}
\tcblower
{\captionsetup{hypcap=false}
\captionof{figure}{Compact rewind-derived optimization record for \texttt{eye}. The highlighted fields show how a code difference becomes reusable knowledge: a retrieval family, a causal mechanism, hardware-profile evidence, transfer constraints, and a short diff excerpt identifying the redundant scattered zero writes.}
\label{fig:appendix_optimization_thought}}
\end{tcolorbox}

Figure~\ref{fig:appendix_optimization_thought} also clarifies why rewind is useful for knowledge construction. The degraded version adds scattered writes to non-diagonal entries; profiling verifies that this edit destroys cache locality and dominates latency. Reversing the validated degradation gives a concrete optimization rule that can be retrieved for later Eye-like kernels, together with explicit conditions under which the rule should not be applied.

\section{Limitations and Broader Impacts}
\label{app:limitations_impact}

\noindent\textbf{Limitations.}
The results apply to existing, functionally correct AscendC implementations on Ascend 910B2 with CANN 8.3. Performance still depends on reusable structure in the starting code, and broader transfer across chips, CANN versions, repositories, and non-transductive benchmark splits remains future work. We also account for hardware-evaluation budget, but finer-grained wall-clock and deployment-cost analysis can be expanded.

\noindent\textbf{Broader Impacts.}
AscendOptimizer may reduce the cost of optimizing operators on knowledge-scarce accelerator platforms, improving hardware utilization and potentially lowering latency and energy use for fixed workloads. Its main risks are that automatically optimized low-level kernels may be incorrect if validation is incomplete, and that measured speedups may not transfer across hardware or software stacks. Optimized operators should therefore be deployed only after task-specific correctness tests and profiling on the target system.

\end{document}